\documentclass{article}

\usepackage[preprint,nonatbib]{neurips_data_2024}

\usepackage[utf8]{inputenc} 
\usepackage[T1]{fontenc}    
\usepackage{url}            
\usepackage{booktabs}       
\usepackage{amsfonts}       
\usepackage{nicefrac}       
\usepackage{microtype}      
\usepackage{times}
\usepackage{epsfig}
\usepackage{graphicx}
\usepackage{amsmath}
\usepackage{amssymb}
\usepackage{booktabs}
\usepackage{comment}
\usepackage{colortbl}
\usepackage{multirow}
\usepackage{eso-pic}
\usepackage[table]{xcolor}
\usepackage{xspace}
\usepackage{xr}
\usepackage[colorlinks,
            linkcolor=blue,
            anchorcolor=blue,
            citecolor=blue,
            ]{hyperref}
\usepackage{comment}

\title{
Open-vocabulary vs. Closed-set: \\Best Practice for Few-shot Object Detection Considering Text Describability
}

\author{%
  Yusuke Hosoya${}^\text{1}$
\quad
Masanori Suganuma${}^\text{1}$
\quad
Takayuki Okatani${}^\text{1,2}$\\
${}^\text{1}$Graduate School of Information Sciences, Tohoku University
\quad
${}^\text{2}$RIKEN Center for AIP\\
{\tt\small \{yhosoya,suganuma,okatani\}@vision.is.tohoku.ac.jp}
}

\makeatletter
\newcommand{\figcaption}[1]{\def\@captype{figure}\caption{#1}}
\newcommand{\tblcaption}[1]{\def\@captype{table}\caption{#1}}
\makeatother

\newcommand{\etal}{\textit{et al.} }

\begin{document}

\maketitle
\begin{abstract}

Open-vocabulary object detection (OVD), detecting specific classes of objects using only their linguistic descriptions (e.g., class names) without any image samples, has garnered significant attention. However, in real-world applications, the target class concepts is often hard to describe in text and the only way to specify target objects is to provide their image examples, yet it is often challenging to obtain a good number of samples. Thus, there is a high demand from practitioners for few-shot object detection (FSOD). A natural question arises: Can the benefits of OVD extend to FSOD for object classes that are difficult to describe in text? Compared to traditional methods that learn only predefined classes (referred to in this paper as closed-set object detection, COD), can the extra cost of OVD be justified? To answer these questions, we propose a method to quantify the ``text-describability'' of object detection datasets using the zero-shot image classification accuracy with CLIP. This allows us to categorize various OD datasets with different text-describability and emprically evaluate the FSOD performance of OVD and COD methods within each category. Our findings reveal that: i) there is little difference between OVD and COD for object classes with low text-describability under equal conditions in OD pretraining; and ii) although OVD can learn from more diverse data than OD-specific data, thereby increasing the volume of training data, it can be counterproductive for classes with low-text-describability. These findings provide practitioners with valuable guidance amidst the recent advancements of OVD methods.
\end{abstract}

\section{Introduction}

Object detection plays a central role in research field of computer vision with a wide range of real-world applications \cite{SSD,FasterRCNN,focalloss,FCOS,DyHead,CenterNet2,DETR,DDETR,DINO,ViTDet}.
Historically, the problem is considered within a closed-set setting, where detectors are designed to identify only the predefined categories of objects encountered during the training process. Recently, the interest in open-vocabulary object detection (OVD) has been growing significantly. Leveraging large models \cite{BERT,RoBERTa,CLIP,ALIGN} that have learned a large amount of text or image-text pairs, it allows for the detection of specific classes of objects based solely on their linguistic descriptions (e.g., class names) without the need for image samples, making it ``zero-shot’’ \cite{OVD_Cap,GLIP,GroundingDINO,Detic,CLIPSelf,MultiModalOVD,TRex2}.

However, in real-world applications of object detection, there are often scenarios where the target classes are difficult to describe with words, such as various types of anomalies in industrial anomaly detection, or lesions that are difficult to identify in medical images. In these cases, it is only possible to specify the target objects by showing image examples, presenting a problem not directly addressed by OVD.

In reality, there is often the additional challenge of not having enough samples available; if sufficient samples were available, the standard supervised learning would work well. Therefore, there are high expectations from practitioners for few-shot object detection (few-shot OD), which can learn to detect objects from only a few examples. 

While various approaches have been tried so far, the best approach to FSOD to date is a rather mediocre one that relies on transfer learning, where a detector pre-trained on some OD data is finetuned with a few-shot examples of the target objects \cite{TFA,FSCE,DeFRCN}. This applies to the traditional OD in the closed setting as well as OVD; while OVD is originally designed for zero-shot detection, existing studies have also attempted to apply their OVD methods to few-shot OD settings, where the same finetuning is the standard \cite{GLIP,GroundingDINO,FoundationFSOD,MQDet}. It should be noted that recent studies have tried to extend OVD to deal with visual prompts— examples to convey concepts that are hard to describe with words \cite{TRex2,DE-ViT,MultiModalOVD}, but broadly speaking, this can be considered a type of few-shot OD.

Considering the above demands for FSOD and the recent advancements of OVD, a natural question that arises is {\em whether the benefits of OVD extend to few-shot OD for object classes that are difficult to describe with words.} Is it superior enough to justify the higher computational costs compared to traditional object detection methods that only learn predetermined classes (referred to as closed-set OD, or COD, in this paper)? What specific advantages do OVD methods offer, which are characterized by similarity calculations in the feature space enabling open-set recognition, the introduction of knowledge from large models (like BERT \cite{BERT} or CLIP \cite{CLIP}), and the increased volume and variety of training data they enable?

To answer these questions, it is essential to understand the difficulty of describing object classes in text. In this paper, we propose a method to quantify the ``text-describability'' of OD datasets based on the zero-shot image classification accuracy of target object classes using CLIP. Using this method, we categorize various OD datasets by their text-describability; see Fig.~\ref{fig:clip_spectrum}. We then experimentally evaluate the performance of OVD and COD methods in FSOD across the introduced dataset categories. 

The results of our experiments show that while OVD significantly outperforms COD under few-shot conditions for easily text-describable classes as expected, there is little difference between the two for classes that are hard to describe in text. Moreover, while OVD can learn from more diverse data, its utility is significant for easily describable classes but can be counterproductive for harder-to-describe classes. These findings are expected to provide some guidance to practitioners amidst the recent advances in various OD methods.

\section{Related Works}\label{sec:related_works}

\subsection{Open-vocabulary Object Detection}

Open-vocabulary object detection (OVD) is an emerging framework for object detection \cite{OVD_Cap, OVD_Distill, RegionCLIP, GLIP, GroundingDINO, DetCLIPv2, Detic, FVLM, CLIPSelf} that has seen significant progress in recent years. Unlike traditional methods (i.e., closed-set object detection (COD)), which can only identify predefined object categories \cite{FasterRCNN, DETR, DyHead}, OVD allows the detection of objects not seen during training. This is achieved using linguistic knowledge from large models such as BERT \cite{BERT} and CLIP \cite{CLIP}. 
To facilitate this capability, existing methods establish a shared feature space between vision and language modalities. They achieve this either by distilling outputs from text encoders \cite{OVD_Cap, OVD_Distill} or by applying text embeddings from pre-trained vision-language models (VLMs) to the classification weights for each category \cite{Detic, OWLViT, FVLM, CLIPSelf}.

\subsection{Few-shot Object Detection}

As it is often difficult to acquire large volumes of training data for object detection \cite{MSCOCO,Object365,LVIS,OpenImages}, training a detector with only a few examples of target objects, known as few-shot object detection (FSOD) \cite{FeatReweight,MetaRCNN,MetaDet,TFA,DeFRCN,FSCE}, has garnered considerable attention. Existing methods for FSOD can be categorized into two approaches: meta-learning  \cite{FeatReweight,MetaRCNN,MetaDet,FSIW} and finetuning \cite{TFA,MSPR,DeFRCN,DCFS,FSCE}. The former approach originally attempts to acquire a ``meta-skill’’ to detect new object classes from only a few samples through the learning of base classes. The latter approach simply involves pre-training on base classes and subsequently training on novel classes, expecting the usual benefits of transfer learning. Recent studies have reported that the finetuning-based approach outperforms the meta-learning-based despite its simplicity \cite{TFA,FSCE,DeFRCN}. Additionally, Wang \etal reported that freezing model parameters except for the final task-specific heads yielded improvements \cite{TFA}. Sun \etal improved this frozen-based approach by employing cosine similarity as classification scores and further added contrastive loss for a RoI head \cite{FSCE}.

FSOD has primarily been studied within the framework of COD. However, in the research of OVD, it has become a norm to report the FSOD performance of OVD methods, in addition to their primary application in zero-shot scenarios \cite{GLIP, GroundingDINO}. In this context, utilizing both textual information and few-shot labeled image examples is expected to improve performance compared to using either one alone. The above insight gained from FSOD in COD seems also applicable to FSOD in OVD. In fact, existing research has shown that finetuning with few-shot examples (where all models, including the text encoder, are subject to training) has become the standard method.

\subsection{Recent FSOD Benchmarks}

Existing FSOD has historically repurposed popular datasets like VOC \cite{VOC} and COCO \cite{MSCOCO} as its benchmarks \cite{FeatReweight,MetaDet,MetaRCNN,TFA,DeFRCN,FSCE}, dividing them into disjoint two splits: base categories and novel categories.
Specifically, PASCAL VOC is partitioned into 15 base and 5 novel categories, while COCO is divided into 60 base and 20 novel categories. 
Whereas these are well-maintained and useful benchmarks, the base and novel categories are sampled from the same dataset, which may be inadequate for evaluating model behaviors in real-world applications with varied target domains.
To explore FSOD effectiveness in more diverse scenarios, recent studies have developed Cross-Domain FSOD (CD-FSOD), assessing performance across multiple image domains \cite{MultiDomainFSOD, Distill-CDFSOD}\footnote{While Lee \etal \cite{MultiDomainFSOD} introduced a similar concept and called it as Multi-domain Few-shot Object Detection (MoFSOD), we consider it identical to CD-FSOD.}. 
Lee \etal \cite{MultiDomainFSOD} and Xiong \etal \cite{Distill-CDFSOD} compiled 10 and 3 datasets from different image domains, respectively, evaluating state-of-the-art FSOD methods. They reported traditional FSOD approaches \cite{TFA,FSCE,DeFRCN} underperformed in the domains distinct from their base category training, highlighting the importance of diverse domain benchmarks.
Their studies provided detailed evaluations using various detectors, but OVD were not investigated.

\begin{figure}[t]
\centering
\includegraphics[width=0.95\linewidth]{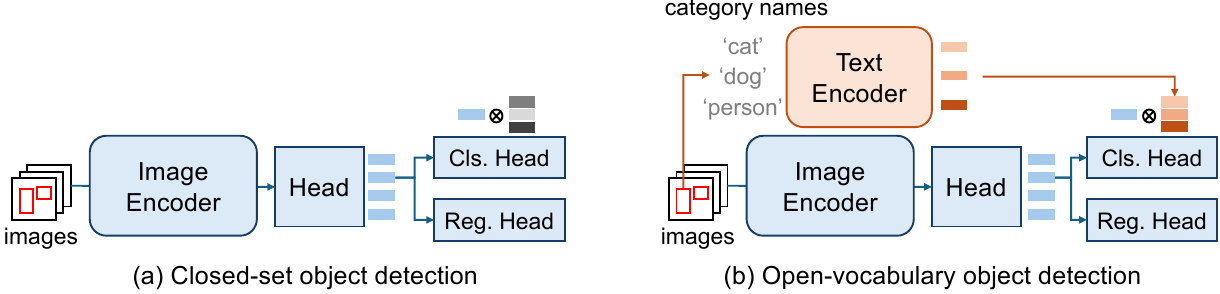}
\caption{An overview of model architectures for (a) closed-set object detection (COD) and (b) open-vocabulary object detection (OVD).}
\label{fig:cod_ovd_abstruct}
\end{figure}

\section{Exploring Best Practice for Few-shot Object Detection}

\subsection{Closed-set and Open-vocabulary Object Detection}

The conventional approach to object detection, referred to as closed-set object detection (COD), operates in the setting  where detectors are trained to identify only predefined object categories present in the training data \cite{FasterRCNN,FCOS,DETR,DyHead,DINO,ViTDet}. 
Figure~\ref{fig:cod_ovd_abstruct}(a) illustrates the model architecture for COD, which features a trainable layer as the final classification head, with dimensions corresponding to the number of target categories.

In contrast, open-vocabulary object detection (OVD) \cite{OVD_Cap, OVD_Distill, Detic, GLIP, OWLViT, DetCLIPv2, CLIPSelf} operates in an open-set setting, leveraging a text encoder, usually derived from pre-trained large models such as BERT \cite{BERT} or CLIP \cite{CLIP}. Figure~\ref{fig:cod_ovd_abstruct}(b)  depicts the general architecture of OVD methods. OVD is characterized by the similarity calculation at the classification head, where text and image features are compared, facilitating open-set recognition. This structure enables the incorporation of textual knowledge into detection and increases the volume and variety of training data, as the models can be trained with more general datasets, such as image-caption pairs \cite{CC3M,SBU}, rather than data specifically designed for object detection.

\subsection{Limitations with Existing OVD Benchmarks}

We are investigating which is more suitable for FSOD between COD and OVD, particularly in cases where object categories are difficult to describe in text and the only option is to present image examples—situations where OVD may not have a significant advantage. If there is an advantage, we expect it to stem from one or more of the three characteristics of OVD mentioned earlier. These questions are critical for practitioners tackling real-world FSOD problems, especially given the recent surge in OVD research.

It is important to note that existing research on OVD has already reported on the performance of FSOD \cite{GLIP,GroundingDINO}. However, these studies do not include comparisons of COD and OVD under the same conditions. More importantly, there is an issue with how datasets for training and testing are selected in current OVD studies, which is crucial for addressing the questions above.

OVD is characterized by pre-training on web-scale data, such as by using BERT or CLIP. In such cases, preventing train-test leakage for common object categories frequently found on the web is extremely difficult. This means that the object categories for which zero-shot/few-shot performance is being tested may have already been pre-trained. As a result, existing OVD research often does not avoid leakage and takes the stance that if the ``dataset'' is different—even if the same object class is being trained—it meets the zero-shot/few-shot requirements. Although this may seem counterintuitive, it is acceptable (or even advantageous) if the goal is to deploy detectors in scenarios with similar image domains and object categories as the training data; the aim is to create a detector that can identify any object as long as it is named.

However, we are focused on detecting object classes that are hard to describe and are necessarily rare on the web, either because the images themselves are rare or because they are not linked to useful text information. This means there is little to no leakage between train and test. Consequently, the few-shot performance for easily describable object categories reported in existing research is likely not useful for predicting the performance of the same detectors under our conditions of interest—where object categories are difficult to describe and there is no leakage between train and test. In other words, we cannot answer the aforementioned questions with the results of existing research.

\subsection{
Categorizing Datasets with Their Text Describability
}\label{sec:spectrum}

To address the aforementioned limitations, it is essential to assess how easily the object classes in an individual object detection dataset\footnote{To be precise, it is more about the tasks, i.e., the target object class list. For clarity, we refer to them as datasets here.} can be described by text. Only then can we explore the relationship between detector performance and the text-describability of the object classes. For example, we can experimentally determine which OVD or COD methods perform better on datasets that are challenging to describe in text.

How can we measure the text-describability of a single dataset? We propose using the zero-shot performance of CLIP as a ``proxy indicator.'' This method involves preparing a collection of datasets $\{D_i\}_{i=1\ldots n}$ and calculating the zero-shot classification accuracy for each dataset $D_i$ in a comparable manner, thereby relativizing its text-describability.

Specifically, for the image input to CLIP, we use the image regions specified by the ground truth bounding boxes for each object class provided by the respective datasets. For text input, we first extract the list $C_i$ of object class names from each dataset $D_i$, create their union $\cap C_i$, consolidate duplicates, and compile a class list $C$ spanning all datasets. We then use prompts (e.g., ``an image of \{class name\}'') based on these classes as text inputs. For each dataset $D_i$, we perform classification on the common class set $C$ using CLIP. The average classification accuracy $a_i$ on the dataset-specific classes $C_i$ (treating classifications into classes in $C\setminus C_i$ as errors) is used as the verbalizability indicator for $D_i$. Considering a classification problem on the common class set $C$ aims to provide a comparable indicator even among datasets with different class counts.

\begin{figure}[t]
\centering
\includegraphics[width=0.95\linewidth]{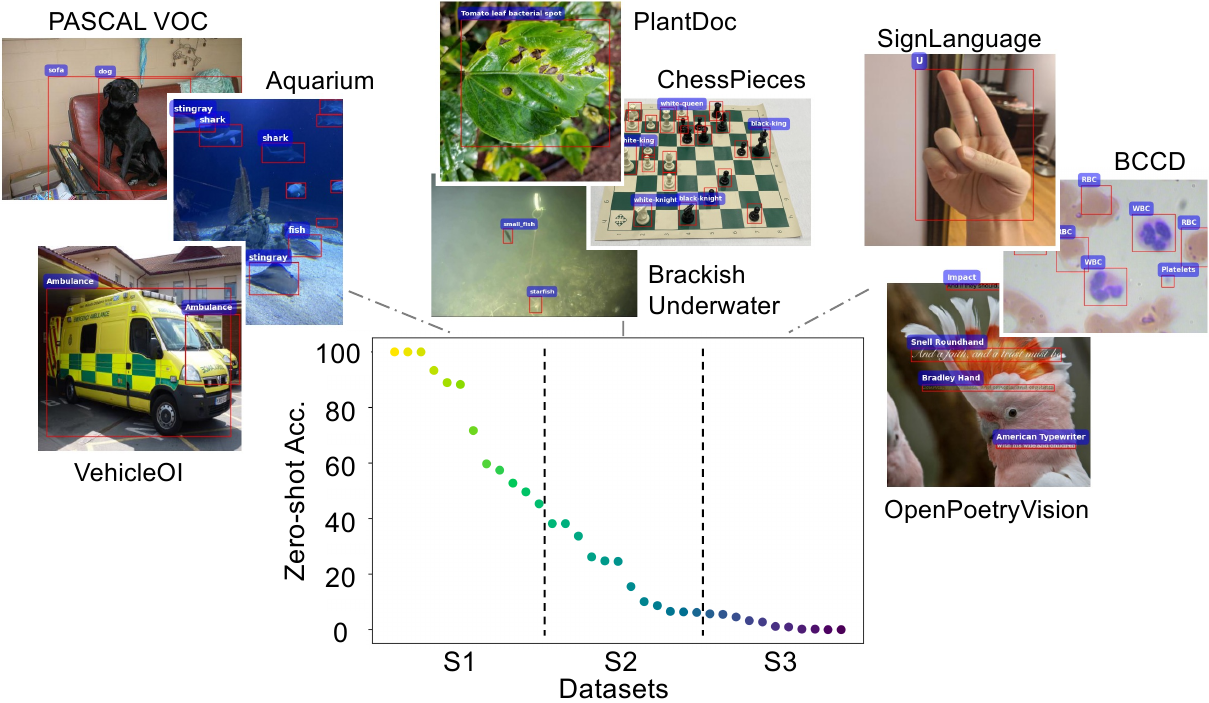}
\caption{Datasets (35 in total from ODinW \cite{Elevater}) sorted by our metric for the difficulty of describing object classes in text. The datasets are categorized and ranked from S1 to S3, indicating decreasing text-describability. }
\label{fig:clip_spectrum}
\end{figure}

In our experiments, we use ODinW (Object Detection in the Wild) \cite{Elevater} for dataset collection, which is a standard approach in recent OVD research \cite{GLIP,GroundingDINO}\footnote{Existing OVD research typically selects 13 out of 35 datasets and uses the average detection accuracy on these to compare methods. Most of these datasets belong to S1 and S2 in our classification, indicating they are easily verbalizable and do not effectively measure performance on less verbalizable datasets.}. This collection includes 35 diverse datasets selected from the 100 available in Roboflow \cite{roboflow100}, each of which simulates a distinct real-world application of object detection. 

Figure~\ref{fig:clip_spectrum} shows how the 35 datasets are sorted using the proposed CLIP-based measure. For statistical evaluation of the detectors' performance, we divided the 35 datasets into three splits (12/12/11 each), labeled S1, S2, and S3, as detailed in the supplementary material. The datasets in S1, S2, and S3 exhibit decreasing CLIP performance, indicating they become less text-describable. As shown in Fig.~\ref{fig:clip_spectrum}, Split S1 includes datasets with common objects, such as the 20 categories of PASCAL VOC \cite{VOC} and common vehicle categories in Open Images \cite{OpenImages}. Split S2 comprises datasets with lower CLIP performance, such as aquatic life in underwater images and fine-grained plant diseases. Split S3 contains datasets like blood cell detection in medical images and sign language detection represented by alphabetical strings.

\paragraph{Remark}
It should be noted that CLIP's zero-shot performance does not directly correspond to the difficulty of verbalizing target objects. The significant variation in CLIP's zero-shot accuracy across different image classification datasets, as reported in the original paper \cite{CLIP}, is likely due to whether the target object classes are included in CLIP's training data. In other words, CLIP's zero-shot performance depends on the abundance of image and class name text pairs in its training data. 

CLIP's training data is widely collected from the web. When data for a particular object class is scarce, there can be two reasons: either the object is difficult to describe, making image-text pairs less likely to exist, or the images themselves are rare due to their specialized domain. Thus, CLIP's performance indicators may combine the difficulty of verbalization and the rarity of images, resulting in only a partial correlation with verbalizability.

However, considering our objective, this might be acceptable. We are interested in how OVD methods perform on data types they have not pre-trained on. Since CLIP's training data is broadly sourced from the web, the training data for OVD should be similar to some extent. Therefore, despite the aforementioned issues, we believe that linking CLIP's performance to the evaluation of OVD methods' performance is useful. Further analyses will be left for future study.

\section{Experiments}\label{sec:experiments}

To answer the above questions, we experimentally evaluate several representative OVD and COD methods in the standard few-shot setting. To ensure the reproducibility of our results, we will make all the code used in our experiments publicly available; see the supplementary material.

\subsection{Compared Methods}\label{sec:compared_methods}

\paragraph{Base Detectors}
We consider four state-of-the-art object detectors: two designed for closed-set object detection (COD)—Dynamic Head (DyHead) \cite{DyHead} and Faster RCNN \cite{FasterRCNN}, and two for open-vocabulary object detection (OVD)—GLIP(A) \cite{GLIP} and F-ViT \cite{CLIPSelf}. DyHead \cite{DyHead} and Faster RCNN \cite{FasterRCNN} are simple yet effective methods for COD, representing one-stage and two-stage detectors, respectively. 
We use Swin-T \cite{SwinT} with Feature Pyramid Network (FPN) \cite{FPN} as their backbones. 

GLIP(A) \cite{GLIP} is an open-vocabulary detector based on DyHead.
It leverages BERT \cite{BERT} as a pre-trained text encoder, to employ its text embeddings as the classification head of the detector. Following the original paper \cite{GLIP}, we utilize Swin-T with FPN as the image encoder.
In Sec.~\ref{sec:results:pretrain_data_amount}, we additionally evaluate GLIP, built on the GLIP(A) architecture but with two modifications over GLIP(A).  
1) GLIP is pre-trained on a more extensive data that includes resources for phrase grounding (GoldG \cite{MDETR}) and image-caption pairs (CC \cite{CC3M} and SBU \cite{SBU}). 2) GLIP incorporates deep fusion modules to enhance the integration of image and text information through cross-attention. These enhancements expectedly expand the vocabulary of visual concepts and allow the model to learn visual features more effectively conditioned on text inputs, both leading to improved OVD performance. 
 
F-ViT \cite{CLIPSelf} is an open-vocabulary detector based on Faster RCNN, using frozen CLIP \cite{CLIP,ViT} both for the image and text encoders.
Before being frozen, the image encoder employs contrastive learning to align dense features of local regions with global features of corresponding crop images. This enables tailored region-level representations for object detection tasks, improving the use of pre-trained CLIP.
Following the original paper \cite{CLIPSelf}, we use EVA-CLIP \cite{EVA-CLIP} for the image and text encoders.

\paragraph{Methods for FSOD Finetuning} Fully finetuning all trainable layers (Full-FT) serves as a baseline in many FSOD studies \cite{MetaRCNN,TFA,DeFRCN,FSCE}. Additionally, we evaluate two state-of-the-art finetuning approaches for FSOD: TFA \cite{TFA} and FSCE \cite{FSCE}. TFA (Two-stage Fine-tuning Approach) \cite{TFA} initially trains all parameters on pre-training phase as usual. Subsequently, only the last prediction heads (i.e., the last layers for classification, regression, centerness, and a projection for text embeddings) are finetuned with few training samples, while the remaining parameters are kept frozen.
FSCE (Few-Shot object detection via Contrastive proposals Encoding) \cite{FSCE} builds upon a frozen-based approach similar to TFA. It enhances TFA by 1) unfreezing Region Proposal Network (RPN) and RoI head, 2) increasing the number of proposals in RPN passed to RoI head, 3) using cosine similarity as classification scores, and 4) adding contrastive proposal encoding loss to its prediction head.
We apply FSCE only to Faster RCNN and F-ViT, considering that it is tailored for two-stage detectors as it adjust the number of RPN proposals.

\subsection{Datasets and Evaluation Protocols}\label{sec:datasets}

\paragraph{Object Detection Pre-training} 

Unless stated otherwise, we utilize Object365-V1 (O365) \cite{Object365}, which comprises 0.61M images across 365 general object categories, as the pre-training dataset for all the detectors\footnote{GLIP \cite{GLIP} reported the number of training images for O365 as 0.66M, but the provided dataset links have expired and cannot be verified. We will use a $\dag$ symbol to indicate this in the results below.}. For GLIP(A) and GLIP, we use their publicly available pre-trained weights from the official repository\footnote{\url{https://github.com/microsoft/GLIP}\label{fn:glip_github}}. Note that this pre-training process is distinctly separate from backbone-level training performed in CLIP \cite{CLIP}, BERT \cite{BERT}, etc.

\paragraph{Evaluation of FSOD Performance}

As previously mentioned, we use the ODinW dataset \cite{Elevater}, which consists of 35 individual object detection (OD) datasets, to evaluate the FSOD performance of the above OD methods; see Sec.~\ref{sec:spectrum} for details of ODinW. We report the average precision (AP) for each method over the intersection over union (IoU) range [0.50:0.95], averaged across datasets within each of the three splits—S1, S2, and S3—each characterized by different levels of text-describability.

For the few-shot configuration, we follow a sampling method employed in previous studies \cite{GLIP,FeatReweight}. Specifically, in $K$-shot settings, we randomly sample the target dataset to ensure that there are at least $K$ images containing one or more ground truth bounding boxes for each category. We consider $K=[1,3,5,10]$ settings. In all experiments, we repeat this sampling process five times using different random seeds and report the averaged performance.

\subsection{Results}\label{sec:results}

\subsubsection{Comparison of COD and OVD Methods}\label{sec:results:cod_ovd}

Table \ref{tbl:cod_ovd} shows the performance of the compared four OD methods on the proposed three splits of ODinW, each with varying numbers $K$ of shots. All methods employ the full-FT approach for FSOD.

It is observed that OVD methods (highlighted in the table) significantly outperform COD methods in the S1 and S2 splits. This is consistent for both one-stage methods (i.e., DyHead and GLIP(A)) and two-stage methods (i.e., Faster RCNN and F-ViT). This result is expected, as OVD methods are designed to detect objects described in text in a zero-shot setting, a capability that also benefits the few-shot setting. Although the performance gap between OVD and COD narrows as $K$ increases, OVD methods consistently show superior performance in S1 and S2 with $K=10$.

Another observation is that the performance gap between OVD and COD methods narrows in the S3 split. Figure \ref{fig:plot_cod_ovd} illustrates the AP ratios of an OVD method compared to its counterpart COD method, highlighting this trend. Specifically, it shows that for S3, GLIP(A)'s performance relative to DyHead's drops to around 1.0, indicating nearly equivalent performance; their APs differ by only about 1.0 AP with $K \ge 3$ (e.g., 39.7 vs. 39.2 at $K=3$).

Moreover, Faster RCNN clearly outperforms its counterpart, F-ViT, with $K=10$ in the S2 split and with $K \ge 3$ in the S3 split. Recall that the datasets in S3 are characterized by low text-describability, such as sign language detection and OCR tasks to identify font names. On these datasets, the superiority of the OVD methods seen in S1 and S2 diminishes. In fact, the COD methods perform even better by a noticeable margin.

\begin{table*}[t]
\footnotesize\centering
\caption{
Few-shot OD performance of COD (closed-set object detection) and OVD (open-vocabulary object detection) methods on the S1, S2, and S3 splits of the 35 ODinW datasets with different numbers $K$ of shots. The values represent the average precision, averaged over the datasets within each split. OVD methods are shaded in gray; IE and TE represent image encoder and text encoder, respectively. }
\label{tbl:cod_ovd}
\scalebox{0.8}{
\renewcommand{\arraystretch}{1.5}
\begin{tabular}{c||cc||ccc|ccc}\specialrule{1.0pt}{0pt}{0pt}
\multirow{2}{*}{Method} & \multicolumn{2}{c||}{Backbone (\#param.)} & \multicolumn{3}{c}{$K=1$} & \multicolumn{3}{|c}{$K=3$} \\ \cline{2-9}
& IE & TE & S1    & S2    & S3    & S1    & S2    & S3    \\ \hline
DyHead & Swin-T(28M) & - & 29.0 $\scriptstyle{\pm 0.8}$ & 22.2 $\scriptstyle{\pm 0.6}$ & 23.8 $\scriptstyle{\pm 1.1}$ & 39.2 $\scriptstyle{\pm 1.6}$ & 33.9 $\scriptstyle{\pm 1.4}$ & 39.7 $\scriptstyle{\pm 0.8}$ \\
\rowcolor{gray!30}GLIP(A) & Swin-T(28M) & BERT(110M) & 37.4 $\scriptstyle{\pm 1.7}$ & 28.5 $\scriptstyle{\pm 0.8}$ & 25.6 $\scriptstyle{\pm 1.3}$ & 44.6 $\scriptstyle{\pm 0.7}$ & 37.1 $\scriptstyle{\pm 0.7}$ & 39.2 $\scriptstyle{\pm 0.5}$ \\
Faster RCNN & Swin-T(28M) & - & 21.7 $\scriptstyle{\pm 2.7}$ & 19.8 $\scriptstyle{\pm 1.1}$ & 21.9 $\scriptstyle{\pm 1.1}$ & 36.2 $\scriptstyle{\pm 1.6}$ & 31.4 $\scriptstyle{\pm 1.2}$ & 38.1 $\scriptstyle{\pm 0.8}$ \\
\rowcolor{gray!30}F-ViT & CLIP-ViT-B/16(86M) & CLIP(63M) & 40.1 $\scriptstyle{\pm 1.1}$ & 24.6 $\scriptstyle{\pm 0.9}$ & 22.9 $\scriptstyle{\pm 1.3}$ & 45.5 $\scriptstyle{\pm 2.5}$ & 32.9 $\scriptstyle{\pm 0.8}$ & 32.0 $\scriptstyle{\pm 0.9}$ \\
\hline\hline

\multirow{2}{*}{Method} & \multicolumn{2}{c||}{Backbone (\#param.)} & \multicolumn{3}{c}{$K=5$} & \multicolumn{3}{|c}{$K=10$} \\ \cline{2-9}
& IE & TE & S1    & S2    & S3    & S1    & S2    & S3    \\ \hline
DyHead & Swin-T(28M) & - & 42.5 $\scriptstyle{\pm 1.6}$ & 36.3 $\scriptstyle{\pm 1.1}$ & 42.9 $\scriptstyle{\pm 1.7}$ & 48.1 $\scriptstyle{\pm 1.2}$ & 41.2 $\scriptstyle{\pm 1.5}$ & 48.7 $\scriptstyle{\pm 1.1}$ \\
\rowcolor{gray!30}GLIP(A) & Swin-T(28M) & BERT(110M) & 49.0 $\scriptstyle{\pm 0.5}$ & 40.2 $\scriptstyle{\pm 0.7}$ & 43.6 $\scriptstyle{\pm 0.7}$ & 52.3 $\scriptstyle{\pm 1.1}$ & 44.5 $\scriptstyle{\pm 1.0}$ & 49.9 $\scriptstyle{\pm 0.7}$ \\
Faster RCNN & Swin-T(28M) & - & 40.1 $\scriptstyle{\pm 2.0}$ & 36.0 $\scriptstyle{\pm 0.6}$ & 42.8 $\scriptstyle{\pm 0.5}$ & 45.7 $\scriptstyle{\pm 1.0}$ & 39.9 $\scriptstyle{\pm 0.9}$ & 48.9 $\scriptstyle{\pm 1.7}$ \\
\rowcolor{gray!30}F-ViT & CLIP-ViT-B/16(86M) & CLIP(63M) & 47.7 $\scriptstyle{\pm 2.6}$ & 36.6 $\scriptstyle{\pm 1.4}$ & 35.2 $\scriptstyle{\pm 1.3}$ & 49.6 $\scriptstyle{\pm 1.5}$ & 40.2 $\scriptstyle{\pm 0.9}$ & 38.7 $\scriptstyle{\pm 1.2}$ \\
\specialrule{0.8pt}{0pt}{0pt}\hline
\end{tabular}
}
\end{table*}

\subsubsection{Impact of Few-shot Finetuning Methods}\label{sec:results:finetuning}

\begin{figure}[t]
    \begin{minipage}[h]{.4\textwidth}
        \centering
        \includegraphics[width=0.9\linewidth]{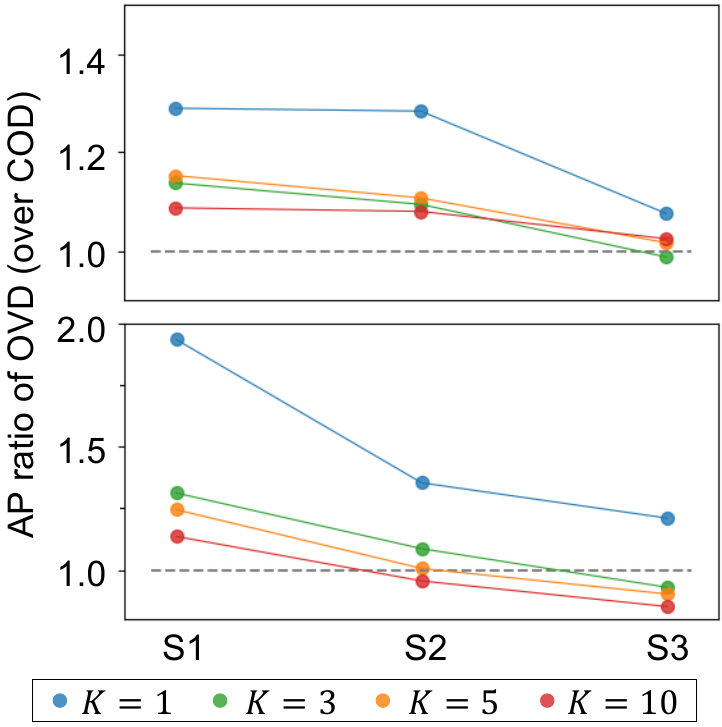}
        \figcaption{AP ratio of OVD/COD. DyHead vs. GLIP(A) (top) and Faster RCNN vs. F-ViT (bottom).}
        \label{fig:plot_cod_ovd}
    \end{minipage}
    \hfill
    \begin{minipage}[h]{.6\textwidth}
    \begin{center}
    \tblcaption{Detection accuracy of state-of-the-art finetuning approaches for FSOD. Results on a $K=3$ are shown. OVD methods are shaded in gray.}
    \label{tbl:ft_approach}
    \scalebox{0.8}{
    \renewcommand{\arraystretch}{1.5}
    \begin{tabular}{c||c|ccc}\specialrule{0.8pt}{0pt}{0pt}\hline 
    Method & \multicolumn{1}{c|}{Finetuning} & S1 & S2 & S3 \\ \hline \hline
    \multirow{2}{*}{DyHead} & Full-FT & 39.2 $\scriptstyle{\pm 1.6}$ & 33.9 $\scriptstyle{\pm 1.4}$ & 39.7 $\scriptstyle{\pm 0.8}$ \\
           & TFA \cite{TFA} & 36.3 $\scriptstyle{\pm 0.7}$ & 23.1 $\scriptstyle{\pm 1.0}$ & 16.0 $\scriptstyle{\pm 0.7}$ \\ \hline
    \rowcolor{gray!30} & Full-FT & 44.6 $\scriptstyle{\pm 0.7}$ & 37.1 $\scriptstyle{\pm 0.7}$ & 39.2 $\scriptstyle{\pm 0.5}$ \\
    \rowcolor{gray!30} \multirow{-2}{*}{GLIP(A)} & TFA \cite{TFA} & 34.5 $\scriptstyle{\pm 0.5}$ & 19.2 $\scriptstyle{\pm 0.6}$ & 10.7 $\scriptstyle{\pm 0.5}$ \\ \hline
    \multirow{3}{*}{Faster RCNN} & Full-FT & 36.2 $\scriptstyle{\pm 1.6}$ & 31.4 $\scriptstyle{\pm 1.2}$ & 38.1 $\scriptstyle{\pm 0.8}$ \\
         & TFA \cite{TFA} & 29.9 $\scriptstyle{\pm 1.3}$ & 22.0 $\scriptstyle{\pm 1.1}$ & 15.4 $\scriptstyle{\pm 0.9}$ \\
         & FSCE \cite{FSCE} & 36.7 $\scriptstyle{\pm 0.6}$ & 28.1 $\scriptstyle{\pm 2.4}$ & 30.5 $\scriptstyle{\pm 1.9}$ \\ \hline
    \rowcolor{gray!30} & Full-FT & 45.5 $\scriptstyle{\pm 2.5}$ & 32.9 $\scriptstyle{\pm 0.8}$ & 32.0 $\scriptstyle{\pm 0.9}$ \\ 
         \rowcolor{gray!30} & TFA \cite{TFA} & 23.6 $\scriptstyle{\pm 0.6}$ & 8.8 $\scriptstyle{\pm 0.2}$ & 5.0 $\scriptstyle{\pm 0.1}$ \\
         \rowcolor{gray!30}\multirow{-3}{*}{F-ViT} & FSCE \cite{FSCE} & 44.7 $\scriptstyle{\pm 1.1}$ & 32.4 $\scriptstyle{\pm 1.2}$ & 34.3 $\scriptstyle{\pm 0.4}$ \\ 
    \specialrule{0.8pt}{0pt}{0pt}\hline
    \end{tabular}
    }
    \end{center}
    \end{minipage}
\end{figure}

We next examine the impact of the fine-tuning methods employed for few-shot learning. Table~\ref{tbl:ft_approach} presents the results for $K=3$ using the same four OD methods with different FSOD fine-tuning approaches. It is observed that TFA \cite{TFA} performs the worst regardless of OVD or COD. Notably, its performance gap compared to Full-FT (i.e., fine-tuning all trainable parameters) increases progressively from S1 to S3.

FSCE \cite{FSCE}, applicable to both Faster RCNN and F-ViT, exhibits similar behavior to TFA, except that F-ViT performs better on S3 with FSCE than with Full-FT. These findings suggest that TFA and FSCE, both recent FSOD fine-tuning methods, do not outperform the standard Full-FT. This holds true regardless of whether the method is COD or OOD and the level of text-describability. This result extends the findings of Lee et al.'s study \cite{MultiDomainFSOD} from COD to OVD, showing that fine-tuning only high-layer parameters improves FSOD performance only when the domain gap between train and test datasets is minimal; otherwise, it negatively impacts performance, and fine-tuning all parameters yields the best results.

\subsubsection{Impact of Pre-training Data}\label{sec:results:pretrain_data_amount}

In FSOD, the detector is initially trained on OD tasks, typically using a large OD dataset and then finetuned with few-shot samples for the target OD task. We examined the impact of this pretraining stage on FSOD with different levels of text-describability. 

Specifically, we used DyHead from COD and studied the effects of the amount of pre-training OD data.
We randomly selected 0.10M images from the COCO dataset (0.12M in total) and 0.61M images from the Objects365 (O365) dataset, combining them to create a 0.20M image dataset. We then created scaled subsets by extracting $x\%$ of images from this combined dataset, maintaining a consistent 1:1 image ratio between COCO and O365. DyHead was trained on these subsets, followed by few-shot adaptation on the S1, S2, and S3 subsets.

The results, shown in Table \ref{tbl:data_amount}, indicate that generally, more pre-training data leads to better FSOD performance. However, a closer examination reveals that the effect is more pronounced for S1 and less so for S3. This likely occurs because the overlap (in terms of object categories and image domains) with the pre-training data decreases in the order of S1, S2, and S3. When targeting S3, although more pre-training data is beneficial, the performance gains diminish compared to S1.

OVD has an advantage over COD in that it can utilize more general image-text pair data, not limited to OD-specific data. We have observed that OVD significantly outperforms COD in S1 and S2 (rows 5 and 6 of the table, copied from Table \ref{tbl:cod_ovd}). This performance gap is expected to widen with the inclusion of non-OD data. However, can this advantage be observed in S3 as well?

To answer this question, we expanded the training data for GLIP(A) under the same conditions for FSOD, resulting in a model referred to as GLIP; see Sec.~\ref{sec:compared_methods} for details. The results, shown in row 7 of Table \ref{tbl:data_amount}, indicate improved accuracy in S1 and S2. Since this method is exclusively applicable to OVD, OVD demonstrates a clear advantage over COD here. However, intriguingly, Table \ref{tbl:data_amount} shows that for S3, GLIP performs worse than GLIP(A) and even falls behind DyHead, a COD method. This suggests that it is safer to use COD for datasets with characteristics like S3. This further supports the above conclusion that FSOD on S3 shows no significant difference between OVD and COD, thus not justifying the extra cost of OVD.

\begin{table}[t]
\footnotesize\centering
\caption{
Detection accuracy across varying amounts of pre-training data. All models are finetuned with Full-FT under a $K=3$ setting. G and C represents grounding datasets (GoldG \cite{MDETR}) and image-caption pairs (CC \cite{CC3M} and SBU \cite{SBU}), respectively. OVD methods are shaded in gray. See Sec.~\ref{sec:datasets} for the $\dag$ indicator.}
\label{tbl:data_amount}
\scalebox{0.9}{
\renewcommand{\arraystretch}{1.5}
\begin{tabular}{c||cc|c|c|ccc}\specialrule{0.8pt}{0pt}{0pt}\hline
    \multirow{2}{*}{Method} & \multicolumn{2}{c|}{Backbone (\#param.)} & \multirow{2}{*}{Pre-training} & \multirow{2}{*}{\#Images} & \multirow{2}{*}{S1} & \multirow{2}{*}{S2} & \multirow{2}{*}{S3} \\ \cline{2-3}
     & IE & TE & & & & & \\ \hline \hline
    & & & & 2K (1\%) & 29.6 $\scriptstyle{\pm 1.8}$ & 27.0 $\scriptstyle{\pm 0.6}$ & 31.1 $\scriptstyle{\pm 0.5}$ \\ 
    & & & & 20K (10\%) & 35.1 $\scriptstyle{\pm 1.0}$ & 29.8 $\scriptstyle{\pm 1.3}$ & 33.7 $\scriptstyle{\pm 0.8}$ \\    
    & & & & 0.10M (50\%) & 40.1 $\scriptstyle{\pm 1.4}$ & 33.6 $\scriptstyle{\pm 0.8}$ & 36.6 $\scriptstyle{\pm 0.4}$ \\
    \multirow{-3}{*}{DyHead} & & & \multirow{-4}{*}{COCO+O365} & 0.20M (100\%) & 40.8 $\scriptstyle{\pm 1.0}$ & 34.0 $\scriptstyle{\pm 1.2}$ & 37.9 $\scriptstyle{\pm 0.3}$ \\ \cline{4-8}
    & \multirow{-5}{*}{Swin-T(28M)} & \multirow{-5}{*}{-} & O365 & 0.61M & 39.2 $\scriptstyle{\pm 1.6}$ & 33.9 $\scriptstyle{\pm 1.4}$ & 39.7 $\scriptstyle{\pm 0.8}$ \\ \hline
    \rowcolor{gray!30} GLIP(A) & Swin-T(28M) & BERT(110M) & O365$^{\dag}$ & 0.66M & 44.6 $\scriptstyle{\pm 0.7}$ & 37.1 $\scriptstyle{\pm 0.7}$ & 39.2 $\scriptstyle{\pm 0.5}$ \\ \hline
    \rowcolor{gray!30} GLIP & Swin-T(28M) & BERT(110M) & O365$^{\dag}$+G+C & 5.46M & 50.4 $\scriptstyle{\pm 0.4}$ & 39.6 $\scriptstyle{\pm 1.2}$ & 34.9 $\scriptstyle{\pm 0.6}$ \\ \hline
\specialrule{0.8pt}{0pt}{0pt}\hline
\end{tabular}
}
\end{table}

\section{Summary and Conclusion}

In this paper, we have addressed the problem of few-shot object detection (FSOD), focusing on the comparison between open-vocabulary object detection (OVD) and closed-set object detection (COD). We first proposed a method to quantify the difficulty of describing target object classes in text using zero-shot image classification accuracy with CLIP. This has enabled us to empirically evaluate COD and OVD methods under equal conditions on various datasets with varying levels of text-describability. Our results provide several key findings. Firstly, for datasets with high text-describability, OVD significantly outperforms COD, as expected. However, when the classes are difficult to describe in text, the superiority of OVD diminishes. Additionally, pre-training on a larger amount of data, which is uniquely beneficial for OVD, can be counterproductive for datasets with low text-describability. These results suggest that for FSOD on datasets where object classes are hard to describe in text, COD methods are recommended over OVD methods. This guidance is valuable for practitioners who are navigating the recent advancements in OVD methods and seeking to optimize their FSOD approaches for specific datasets.

{\small
\bibliographystyle{ieee_fullname}
\bibliography{egbib}
}
\clearpage
\appendix

\section*{\Large{Appendix}}

\section{Access to Our Code}\label{sec:sup:access_to_data}
All the code used in our experiments is available at: {\small\url{https://github.com/rsCPSyEu/ovd_cod}}
\section{Detailed Design of Our Benchmark Test}

Our benchmark test is designed by repurposing the ODinW (Object Detection in the Wild) datasets \cite{Elevater}. The 35 included datasets are partitioned into three splits based on their CLIP zero-shot performance, as explained in Sec.~\ref{sec:spectrum}. 

Table ~\ref{tbl:sup:dataset_list} shows the list of datasets included in each split and the CLIP classification accuracy for each dataset. See also Fig.~\ref{fig:clip_spectrum} in the main paper. The datasets highlighted in red indicate the 13 datasets that existing OVD studies employ for evaluation. Most of these belong to S1 and S2, suggesting that their class names are relatively easy to describe in text. S3 consists only of datasets with classes that are difficult to describe. 

\renewcommand{\arraystretch}{1.3}
\begin{table*}[h]
\centering
\caption{A list of 35 Datasets in ODinW \cite{Elevater} with corresponding CLIP zero-shot accuracy. Red color shows 13 datasets that are typically used in previous methods \cite{GLIP}.}
\label{tbl:sup:dataset_list}
\scalebox{0.7}{
\begin{tabular}{cc||cc||cc}\specialrule{0.7pt}{0pt}{0pt}\hline
\multicolumn{2}{c||}{S1(\#12)} & \multicolumn{2}{c||}{S2(\#12)} & \multicolumn{2}{c}{S3(\#11)}  \\ \hline
Dataset & Acc. & Dataset & Acc. & Dataset & Acc. \\ \hline

\begin{tabular}{c} 
{\color{red}CottontailRabbits} \\
{\color{red}Packages} \\
{\color{red}Raccoon} \\ 
{\color{red}NorthAmericaMushrooms} \\
{\color{red}Pothole} \\
OxfordPets(breed) \\
{\color{red}VehiclesOpenImages} \\
MountainDewCommercial \\
{\color{red}PASCAL VOC} \\
OxfordPets(species) \\
WildfireSmoke \\
{\color{red}Aquarium}  \\
\end{tabular} & 

\begin{tabular}{c} 
100.0 \\ 100.0 \\ 100.0 \\ 93.3 \\ 88.9 \\ 87.8 \\ 71.7 \\ 59.7 \\ 57.6 \\ 56.6 \\ 52.7 \\ 49.6
\end{tabular} & 

\begin{tabular}{c} 
{\color{red}ShellfishOpenImages} \\
{\color{red}AerialMaritimeDrone(large)} \\
AerialMaritimeDrone(tiled) \\ 
SelfDrivingCar  \\
Plantdoc \\
BrackishUnderwater  \\
{\color{red}Pistols} \\
MaskWearing \\
{\color{red}EgoHands(generic)} \\
ChessPiecesPieces \\
ThermalCheetah \\
DroneControl  \\
\end{tabular} & 

\begin{tabular}{c} 
45.3 \\ 38.2 \\ 38.2 \\ 33.7 \\ 26.2 \\ 24.7 \\ 24.6 \\ 15.5 \\ 10.0 \\ 8.6 \\ 6.5 \\ 6.3
\end{tabular} & 

\begin{tabular}{c} 
HardHatWorkers \\
AmericanSignLanguageLetters \\
UnoCards \\
BoggleBoards \\
WebsiteScreenshots \\
{\color{red}ThermalDogsAndPeople} \\
BCCD \\
Dice \\
OpenPoetryVision \\
PKLot640 \\
EgoHands(specific) \\
\end{tabular} &

\begin{tabular}{c} 
6.2 \\ 5.7 \\ 5.5 \\ 4.5 \\ 2.8 \\ 1.4 \\ 0.9 \\ 0.2 \\ 0.1 \\ 0.0 \\ 0.0 
\end{tabular} \\

\specialrule{0.7pt}{0pt}{0pt}\hline 
\end{tabular}
}
\end{table*}
\renewcommand{\arraystretch}{1}

In our experiments, we evaluate the object detection performance of multiple methods in a few-shot setting. For each dataset listed in Table ~\ref{tbl:sup:dataset_list}, we randomly select few-shot training samples from the dataset's training split; the original validation/testing split is used for validation/testing. We repeat this sampling, followed by training and evaluating each method five times, and report the average mean AP across all object classes.

We consider a $K$-shot setting where $K = 1$, 3, 5, and 10. Following previous studies, $K$ indicates the number of images per object class, not the number of bounding boxes. Depending on the dataset, a single image may contain multiple object instances annotated with different bounding boxes, so the number of bounding boxes used for training varies.

Table \ref{tbl:sup:data_cfg_k13} and \ref{tbl:sup:data_cfg_k510} provide detailed configurations. For each dataset listed in a row, the column ``classes'' indicates the number of object classes. Columns from ``seed=0'' to ``seed=4'' represent individual trials in the five random samplings, each reporting the number of images and bounding boxes, separated by `/,' used for training.

\renewcommand{\arraystretch}{1.3}
\begin{table*}[t]
\centering
\caption{
Detailed configurations of the few-shot training per a dataset. The numbers separated by a slash represent the total number of images (\#Img.) and the total number of bounding boxes (\#Ann.) used for the few-shot training, respectively. Configurations for $K=1$ and 3 are shown.
}
\label{tbl:sup:data_cfg_k13}
\scalebox{0.65}{
\begin{tabular}{c||c|ccccc|ccccc}
\hline
& & \multicolumn{5}{c|}{$K=1$ ( \#Img. / \#Ann. )} & \multicolumn{5}{c}{$K=3$ ( \#Img. / \#Ann. )} \\ \cline{3-12}
\multirow{-2}{*}{Dataset} & \multirow{-2}{*}{Classes} & seed=0 & seed=1 & seed=2 & seed=3 & seed=4 & seed=0 & seed=1 & seed=2 & seed=3 & seed=4 \\ \hline
CottontailRabbits & 1 & 1/1 & 1/1 & 1/1 & 1/1 & 1/1 & 3/3 & 3/3 & 3/3 & 3/3 & 3/3 \\ 
Packages & 1 & 1/2 & 1/1 & 1/2 & 1/2 & 1/2 & 3/6 & 3/4 & 3/4 & 3/4 & 3/5 \\ 
Raccoon & 1 & 1/1 & 1/1 & 1/1 & 1/1 & 1/1 & 3/4 & 3/3 & 3/3 & 3/3 & 3/3 \\ 
NorthAmericaMushrooms & 2 & 2/8 & 2/2 & 2/4 & 2/3 & 2/2 & 6/14 & 6/7 & 6/9 & 6/15 & 6/7 \\ 
Pothole & 1 & 1/1 & 1/3 & 1/1 & 1/2 & 1/3 & 3/4 & 3/24 & 3/5 & 3/6 & 3/9 \\ 
OxfordPets(breed) & 37 & 37/37 & 37/37 & 37/37 & 37/37 & 37/37 & 111/111 & 111/111 & 111/111 & 111/111 & 111/111 \\ 
VehiclesOpenImages & 5 & 5/14 & 5/18 & 5/7 & 5/17 & 5/6 & 15/32 & 15/46 & 15/25 & 15/28 & 15/28 \\ 
MountainDewCommercial & 1 & 1/39 & 1/49 & 1/51 & 1/51 & 1/24 & 3/90 & 3/110 & 3/98 & 3/90 & 3/78 \\ 
PASCAL VOC & 20 & 17/35 & 17/42 & 18/41 & 19/54 & 16/40 & 50/126 & 51/143 & 51/134 & 55/156 & 51/132 \\ 
OxfordPets(species) & 2 & 2/2 & 2/2 & 2/2 & 2/2 & 2/2 & 6/6 & 6/6 & 6/6 & 6/6 & 6/6 \\ 
WildfireSmoke & 1 & 1/1 & 1/1 & 1/1 & 1/1 & 1/1 & 3/3 & 3/3 & 3/3 & 3/3 & 3/3 \\ 
Aquarium & 7 & 6/50 & 6/53 & 6/49 & 6/26 & 6/15 & 17/151 & 19/100 & 18/107 & 18/97 & 17/90 \\ 
ShellfishOpenImages & 3 & 3/7 & 3/3 & 3/3 & 3/5 & 3/6 & 9/16 & 9/15 & 9/10 & 9/19 & 9/18 \\ 
AerialMaritimeDrone(large) & 5 & 3/96 & 4/53 & 2/48 & 2/28 & 3/36 & 6/141 & 10/139 & 7/110 & 5/76 & 7/109 \\ 
AerialMaritimeDrone(tiled) & 5 & 4/10 & 3/8 & 3/11 & 5/23 & 4/13 & 12/42 & 12/40 & 10/39 & 13/48 & 12/42 \\ 
SelfDrivingCar & 11 & 6/50 & 8/88 & 7/64 & 9/98 & 9/74 & 20/194 & 25/261 & 22/190 & 24/227 & 24/222 \\ 
Plantdoc & 30 & 30/123 & 30/142 & 30/123 & 30/137 & 30/158 & 86/323 & 86/417 & 87/344 & 87/421 & 86/381 \\ 
BrackishUnderwater & 6 & 6/18 & 5/12 & 4/16 & 5/16 & 5/22 & 18/42 & 15/41 & 13/38 & 14/36 & 15/42 \\ 
Pistols & 1 & 1/1 & 1/1 & 1/1 & 1/1 & 1/1 & 3/3 & 3/3 & 3/3 & 3/3 & 3/3 \\ 
MaskWearing & 2 & 1/20 & 2/3 & 2/24 & 2/5 & 2/11 & 5/48 & 5/17 & 4/45 & 5/18 & 6/34 \\ 
EgoHands(generic) & 1 & 1/4 & 1/3 & 1/3 & 1/2 & 1/4 & 3/10 & 3/8 & 3/11 & 3/8 & 3/10 \\ 
ChessPiecesPieces & 13 & 3/52 & 5/73 & 3/31 & 2/30 & 3/34 & 8/131 & 10/135 & 9/128 & 6/106 & 8/130 \\ 
ThermalCheetah & 2 & 2/6 & 2/7 & 2/5 & 2/6 & 2/4 & 6/17 & 6/16 & 6/17 & 6/17 & 6/14 \\ 
DroneControl & 8 & 8/10 & 8/10 & 8/10 & 8/12 & 8/10 & 23/27 & 23/27 & 23/27 & 23/29 & 23/27 \\ 
HardHatWorkers & 3 & 3/11 & 3/16 & 3/14 & 3/10 & 2/17 & 9/39 & 8/39 & 9/38 & 8/29 & 7/35 \\ 
AmericanSignLanguageLetters & 26 & 26/26 & 26/26 & 26/26 & 26/26 & 26/26 & 78/78 & 78/78 & 78/78 & 78/78 & 78/78 \\ 
UnoCards & 15 & 9/27 & 8/24 & 11/33 & 8/24 & 10/30 & 23/69 & 23/69 & 21/63 & 23/69 & 21/63 \\ 
BoggleBoards & 36 & 12/301 & 14/373 & 14/324 & 16/396 & 15/378 & 30/787 & 32/859 & 36/903 & 32/801 & 36/943 \\ 
WebsiteScreenshots & 8 & 2/91 & 4/166 & 4/129 & 3/110 & 3/98 & 7/327 & 10/356 & 11/446 & 11/569 & 10/385 \\ 
ThermalDogsAndPeople & 2 & 2/3 & 2/3 & 1/2 & 2/3 & 2/2 & 6/8 & 5/7 & 4/8 & 6/7 & 6/7 \\ 
BCCD & 3 & 2/34 & 1/16 & 3/28 & 2/29 & 2/22 & 4/62 & 4/56 & 6/73 & 5/78 & 4/56 \\ 
Dice & 6 & 5/8 & 4/6 & 6/12 & 3/45 & 5/8 & 14/23 & 13/58 & 14/25 & 13/59 & 12/64 \\ 
OpenPoetryVision & 43 & 24/85 & 26/75 & 21/82 & 28/104 & 23/79 & 62/208 & 66/212 & 53/182 & 67/229 & 60/202 \\ 
PKLot640 & 2 & 1/40 & 1/100 & 1/100 & 1/28 & 1/100 & 4/280 & 3/240 & 3/168 & 3/228 & 4/268 \\ 
EgoHands(specific) & 4 & 2/6 & 2/6 & 1/4 & 2/7 & 2/6 & 5/17 & 4/14 & 3/12 & 5/17 & 5/15 \\ 
\specialrule{0.7pt}{0pt}{0pt}\hline 
\end{tabular}
}
\end{table*}
\renewcommand{\arraystretch}{1}

\renewcommand{\arraystretch}{1.3}
\begin{table*}[t]
\centering
\caption{
Detailed configurations of the few-shot training per a dataset. The numbers separated by a slash represent the total number of images (\#Img.) and the total number of bounding boxes (\#Ann.) used for the few-shot training, respectively. Configurations for $K=5$ and $10$ are shown.}
\label{tbl:sup:data_cfg_k510}
\scalebox{0.6}{
\begin{tabular}{c||c|ccccc|ccccc}
\hline
& & \multicolumn{5}{c|}{$K=5$ ( \#Img / \#Ann )} & \multicolumn{5}{c}{$K=10$ ( \#Img / \#Ann )} \\ 
\multirow{-2}{*}{Dataset} & \multirow{-2}{*}{Classes} & seed=0 & seed=1 & seed=2 & seed=3 & seed=4 & seed=0 & seed=1 & seed=2 & seed=3 & seed=4 \\ \hline

CottontailRabbits & 1 & 5/5 & 5/5 & 5/5 & 5/5 & 5/5 & 10/10 & 10/11 & 10/10 & 10/10 & 10/11 \\ 
Packages & 1 & 5/9 & 5/7 & 5/7 & 5/8 & 5/7 & 10/18 & 10/16 & 10/16 & 10/18 & 10/15 \\ 
Raccoon & 1 & 5/6 & 5/5 & 5/5 & 5/5 & 5/5 & 10/12 & 10/10 & 10/10 & 10/11 & 10/13 \\ 
NorthAmericaMushrooms & 2 & 10/24 & 10/17 & 10/16 & 10/19 & 10/12 & 20/38 & 20/32 & 20/27 & 20/30 & 20/25 \\ 
Pothole & 1 & 5/15 & 5/31 & 5/10 & 5/8 & 5/12 & 10/29 & 10/46 & 10/22 & 10/29 & 10/26 \\ 
OxfordPets(breed) & 37 & 185/186 & 185/185 & 185/185 & 185/185 & 185/185 & 370/371 & 370/370 & 370/370 & 370/371 & 370/371 \\ 
VehiclesOpenImages & 5 & 25/49 & 25/63 & 25/44 & 25/48 & 25/45 & 50/91 & 50/119 & 50/97 & 50/81 & 50/85 \\ 
MountainDewCommercial & 1 & 5/126 & 5/144 & 5/149 & 5/137 & 5/178 & 10/264 & 10/324 & 10/276 & 10/267 & 10/323 \\ 
PASCAL VOC & 20 & 84/259 & 85/223 & 85/262 & 88/286 & 84/249 & 169/618 & 166/496 & 172/535 & 171/501 & 168/490 \\ 
OxfordPets(species) & 2 & 10/10 & 10/10 & 10/10 & 10/10 & 10/10 & 20/20 & 20/20 & 20/20 & 20/20 & 20/20 \\ 
WildfireSmoke & 1 & 5/5 & 5/5 & 5/5 & 5/5 & 5/5 & 10/10 & 10/10 & 10/10 & 10/10 & 10/10 \\ 
Aquarium & 7 & 31/198 & 32/187 & 29/175 & 32/200 & 29/208 & 60/353 & 65/421 & 60/497 & 64/359 & 61/408 \\ 
ShellfishOpenImages & 3 & 15/28 & 15/25 & 15/24 & 15/28 & 15/42 & 30/59 & 30/60 & 30/56 & 30/56 & 30/75 \\ 
AerialMaritimeDrone(large) & 5 & 10/221 & 15/290 & 11/167 & 10/155 & 11/236 & 24/402 & 27/486 & 25/394 & 22/297 & 23/424 \\ 
AerialMaritimeDrone(tiled) & 5 & 17/53 & 20/68 & 18/65 & 20/71 & 19/64 & 37/122 & 40/128 & 38/136 & 37/132 & 41/124 \\ 
SelfDrivingCar & 11 & 32/336 & 39/399 & 37/363 & 35/315 & 40/409 & 65/667 & 73/698 & 72/701 & 74/660 & 70/716 \\ 
Plantdoc & 30 & 139/519 & 140/629 & 141/576 & 139/601 & 139/655 & 273/981 & 272/1108 & 274/1116 & 274/1119 & 274/1124 \\ 
BrackishUnderwater & 6 & 29/70 & 25/64 & 23/60 & 25/59 & 26/84 & 52/145 & 47/135 & 48/116 & 51/148 & 50/137 \\ 
Pistols & 1 & 5/8 & 5/5 & 5/5 & 5/5 & 5/5 & 10/18 & 10/11 & 10/10 & 10/10 & 10/10 \\ 
MaskWearing & 2 & 8/54 & 8/43 & 7/63 & 9/40 & 9/48 & 16/101 & 15/69 & 15/123 & 15/59 & 16/116 \\ 
EgoHands(generic) & 1 & 5/16 & 5/15 & 5/16 & 5/13 & 5/16 & 10/31 & 10/31 & 10/32 & 10/30 & 10/34 \\ 
ChessPiecesPieces & 13 & 14/194 & 14/230 & 13/235 & 10/191 & 11/190 & 25/315 & 27/401 & 25/403 & 24/358 & 20/366 \\ 
ThermalCheetah & 2 & 10/25 & 10/32 & 10/32 & 10/30 & 10/28 & 18/50 & 18/48 & 18/50 & 17/49 & 18/48 \\ 
DroneControl & 8 & 37/42 & 37/41 & 37/41 & 37/43 & 37/42 & 72/78 & 72/77 & 72/76 & 71/78 & 72/77 \\ 
HardHatWorkers & 3 & 15/75 & 14/67 & 14/59 & 13/47 & 13/71 & 30/156 & 28/157 & 29/140 & 28/144 & 28/155 \\ 
AmericanSignLanguageLetters & 26 & 130/130 & 130/130 & 130/130 & 130/130 & 130/130 & 260/260 & 260/260 & 260/260 & 260/260 & 260/260 \\ 
UnoCards & 15 & 38/114 & 36/108 & 35/105 & 36/108 & 34/102 & 69/207 & 65/195 & 69/207 & 69/207 & 64/192 \\ 
BoggleBoards & 36 & 49/1280 & 51/1383 & 52/1328 & 50/1298 & 57/1448 & 77/1935 & 80/2085 & 80/2003 & 76/1910 & 86/2141 \\ 
WebsiteScreenshots & 8 & 13/610 & 16/723 & 19/725 & 18/734 & 17/827 & 29/1418 & 31/1259 & 37/1565 & 34/1468 & 32/1412 \\ 
ThermalDogsAndPeople & 2 & 9/12 & 9/12 & 8/14 & 9/10 & 9/12 & 17/26 & 16/23 & 16/25 & 18/22 & 17/23 \\ 
BCCD & 3 & 8/107 & 8/107 & 8/100 & 7/110 & 7/84 & 17/251 & 15/194 & 14/180 & 14/209 & 15/197 \\ 
Dice & 6 & 23/68 & 20/83 & 22/38 & 20/102 & 20/79 & 46/106 & 40/155 & 46/73 & 40/138 & 37/149 \\ 
OpenPoetryVision & 43 & 93/305 & 105/330 & 94/307 & 108/359 & 91/300 & 172/555 & 184/563 & 186/572 & 180/576 & 177/564 \\ 
PKLot640 & 2 & 7/364 & 5/440 & 5/248 & 7/424 & 7/424 & 15/972 & 12/1008 & 12/684 & 12/804 & 14/800 \\ 
EgoHands(specific) & 4 & 9/30 & 6/20 & 7/23 & 10/33 & 9/28 & 18/61 & 13/45 & 13/43 & 18/59 & 15/48 \\ 
\specialrule{0.7pt}{0pt}{0pt}\hline 
\end{tabular}
}
\end{table*}
\renewcommand{\arraystretch}{1}

\section{More Details of Experimental Settings}

This section provides the comprehensive configurations of our experiments reported in the paper.

\subsection{Training}

\subsubsection{Object Detection Pre-training}
To pre-train the models with the Object365 dataset, we utilized the AdamW optimizer \cite{AdamW}, setting the batch size to 64 across 8 V100 GPUs. The training duration was 30 epochs for DyHead, GLIP(A), and Faster RCNN, while F-ViT was trained for 20 epochs. Notably, the original F-ViT training protocol \cite{CLIPSelf} employed only 3 epochs, leveraging its robust pre-training via CLIPSelf self-supervised learning.

The initial learning rate was set at $1.0 \times 10^{-4}$. The weight decay parameters were set at 0.1 for F-ViT and 0.05 for the other models. The learning rate was reduced by a factor of 10 at 67\% and 89\% of the total iterations. For data augmentation, we applied a standard random horizontal flip with a 0.5 probability and implemented multi-scale training. Input images were resized such that their shorter side is sampled from [480,560,640,720,800].

\subsubsection{Finetuning}
The finetuning process retains the same training configurations as pre-training but includes several modifications.
For all finetuning datasets (i.e., ODinW), the batch size is set to 4 on 4 V100 GPUs. 
We specifically increase the initial learning rate to $1.0 \times 10^{-3}$ for DyHead with the TFA \cite{TFA} approach.
In line with the GLIP \cite{GLIP} implementation, we adjust learning rates during finetuning, based on the detection performance assessed on each dataset's validation data. Specifically, we employ a Pytorch ReduceLROnPlateau scheduler with a patience of 3 and a factor of 0.1 to decrease the learning rate when no improvement is observed in the validation dataset. Furthermore, we terminate the fine-tuning process if there is no improvement in validation for 8 consecutive epochs.

\section{More Experimental Results}

This section reports all the detection performance results for reference, including those omitted in the main paper due to space constraints.

\subsection{Results of Full-FT}
Table~\ref{tbl:sup:all_results_fullft_dyhead} through \ref{tbl:sup:all_results_fullft_fvit} show all fintuning results across different number of few-shot samples ($K=[1,3,5,10]$) and different random seeds (seed$=[0,1,2,3,4]$) for the compared methods, including DyHead, GLIP(A), GLIP, Faster RCNN, and F-ViT, respectively.  
Each model is finetuned using Full-FT approach.

\subsection{Results of Finetuning Approaches}

In Sec.~\ref{sec:results:finetuning}, we evalute TFA (Two-stage Fine-tuning Approach) \cite{TFA} and FSCE (Few-Shot object detection via Contrastive proposals Encoding) \cite{FSCE} as finetuning approaches.
Table~\ref{tbl:sup:all_results_tfa_dyhead} to \ref{tbl:sup:all_results_tfa_fvit} show the results of TFA with four main compared methods (DyHead, GLIP(A), Faster RCNN, and F-ViT, sequentially). 
Additionally, Table~\ref{tbl:sup:all_results_fsce_frcnn} and \ref{tbl:sup:all_results_fsce_fvit} show the results of FSCE. 
These fine-tuning approaches are evaluated only in the $K=3$ setting.

\subsection{Results of Different Pre-training Data}

Table~\ref{tbl:sup:all_results_data_amount} presents all results associated with the experiments on pre-training data in Sec~\ref{sec:results:pretrain_data_amount} in the main paper. 

\renewcommand{\arraystretch}{1.2}
\begin{table}[t]
\footnotesize\centering
\caption{Detection performance for DyHead. All methods are finetuned with Full-FT approach.}
\label{tbl:sup:all_results_fullft_dyhead}
\scalebox{0.58}{ 
\begin{tabular}{c||ccccc|ccccc|ccccc|ccccc}
\specialrule{0.8pt}{0pt}{0pt}
 & \multicolumn{5}{c|}{K=1} & \multicolumn{5}{c|}{K=3} & \multicolumn{5}{c|}{K=5} & \multicolumn{5}{c}{K=10} \\ 
Seed & 0 & 1 & 2 & 3 & 4 & 0 & 1 & 2 & 3 & 4 & 0 & 1 & 2 & 3 & 4 & 0 & 1 & 2 & 3 & 4 \\ \hline

CottontailRabbits & 63.8 & 44.7 & 51.6 & 58.3 & 60.4 & 62.4 & 54.7 & 57.6 & 56.4 & 54.2 & 70.7 & 66.1 & 54.9 & 62.5 & 62.1 & 67.5 & 66.7 & 53.3 & 66.0 & 65.5 \\ 
Packages & 44.5 & 61.5 & 30.6 & 51.1 & 68.6 & 43.5 & 62.3 & 56.3 & 45.9 & 73.6 & 63.1 & 60.1 & 55.2 & 39.1 & 62.8 & 48.4 & 60.7 & 55.4 & 59.1 & 63.9 \\ 
Raccoon & 41.6 & 33.8 & 32.1 & 34.8 & 32.6 & 30.8 & 35.5 & 43.2 & 42.7 & 34.9 & 33.6 & 27.7 & 50.9 & 43.3 & 50.4 & 44.5 & 46.2 & 54.4 & 45.8 & 27.4 \\ 
NorthAmericaMushrooms & 77.7 & 68.0 & 86.2 & 66.0 & 68.9 & 81.5 & 39.6 & 86.8 & 85.4 & 71.5 & 84.2 & 66.9 & 89.3 & 78.5 & 72.1 & 90.7 & 69.1 & 84.7 & 84.4 & 85.8 \\ 
Pothole & 3.3 & 18.8 & 12.8 & 9.8 & 12.2 & 19.8 & 18.8 & 22.4 & 18.8 & 18.5 & 26.4 & 20.3 & 24.8 & 18.7 & 14.5 & 28.6 & 19.1 & 30.3 & 27.4 & 27.4 \\ 
OxfordPets(breed) & 14.7 & 14.6 & 16.8 & 16.9 & 11.9 & 33.0 & 33.6 & 33.4 & 33.2 & 35.2 & 45.4 & 44.8 & 46.6 & 44.5 & 45.3 & 59.9 & 62.2 & 61.5 & 59.8 & 60.3 \\
\end{tabular}
}
\end{table}
\renewcommand{\arraystretch}{1}


\renewcommand{\arraystretch}{1.2}
\begin{table}[t]
\footnotesize\centering
\caption{Detection performance for GLIP(A). All methods are finetuned with Full-FT approach.}
\label{tbl:sup:all_results_fullft_glipa}
\scalebox{0.55}{ 
\begin{tabular}{c||ccccc|ccccc|ccccc|ccccc}
\specialrule{0.8pt}{0pt}{0pt}
 & \multicolumn{5}{c|}{K=1} & \multicolumn{5}{c|}{K=3} & \multicolumn{5}{c|}{K=5} & \multicolumn{5}{c}{K=10} \\ 
Seed & 0 & 1 & 2 & 3 & 4 & 0 & 1 & 2 & 3 & 4 & 0 & 1 & 2 & 3 & 4 & 0 & 1 & 2 & 3 & 4 \\ \hline

CottontailRabbits & 65.0 & 61.4 & 66.0 & 53.3 & 67.0 & 65.5 & 62.2 & 62.0 & 65.1 & 69.1 & 62.6 & 66.4 & 64.3 & 64.1 & 67.0 & 63.6 & 70.8 & 57.7 & 67.9 & 68.5 \\ 
Packages & 58.6 & 42.8 & 56.9 & 61.2 & 75.2 & 57.9 & 62.6 & 70.3 & 63.0 & 75.6 & 59.1 & 72.2 & 72.8 & 62.0 & 70.2 & 64.3 & 56.9 & 74.3 & 45.6 & 54.3 \\ 
Raccoon & 45.5 & 53.0 & 55.9 & 58.0 & 50.4 & 56.5 & 55.3 & 55.4 & 54.3 & 48.4 & 59.6 & 55.2 & 56.9 & 55.1 & 53.6 & 44.0 & 58.7 & 54.3 & 59.1 & 55.8 \\ 
NorthAmericaMushrooms & 74.8 & 56.5 & 70.4 & 70.0 & 62.1 & 79.9 & 66.1 & 72.4 & 84.9 & 71.2 & 85.4 & 68.2 & 86.5 & 84.4 & 65.9 & 86.3 & 70.7 & 79.4 & 84.4 & 68.0 \\ 
Pothole & 10.3 & 8.2 & 6.6 & 19.2 & 19.1 & 23.7 & 18.0 & 15.9 & 23.3 & 16.9 & 27.2 & 24.3 & 24.5 & 25.5 & 25.5 & 32.3 & 32.2 & 32.9 & 32.8 & 30.8 \\ 
OxfordPets(breed) & 11.8 & 13.3 & 10.0 & 12.5 & 12.0 & 30.3 & 31.3 & 32.7 & 29.9 & 29.9 & 43.7 & 43.8 & 46.1 & 42.6 & 45.0 & 60.4 & 56.2 & 62.7 & 57.2 & 60.9 \\
\end{tabular}
}
\end{table}
\renewcommand{\arraystretch}{1}

\renewcommand{\arraystretch}{1.2}
\begin{table}[t]
\footnotesize\centering
\caption{Detection performance for GLIP. All methods are finetuned with Full-FT approach}
\label{tbl:sup:all_results_fullft_glip}
\scalebox{0.55}{ 
\begin{tabular}{c||ccccc|ccccc|ccccc|ccccc}
\specialrule{0.8pt}{0pt}{0pt}
 & \multicolumn{5}{c|}{K=1} & \multicolumn{5}{c|}{K=3} & \multicolumn{5}{c|}{K=5} & \multicolumn{5}{c}{K=10} \\ 
Seed & 0 & 1 & 2 & 3 & 4 & 0 & 1 & 2 & 3 & 4 & 0 & 1 & 2 & 3 & 4 & 0 & 1 & 2 & 3 & 4 \\ \hline

CottontailRabbits & 69.0 & 69.0 & 70.1 & 69.0 & 70.6 & 66.7 & 63.0 & 69.0 & 71.0 & 70.0 & 69.0 & 71.4 & 67.3 & 71.9 & 70.8 & 72.2 & 73.7 & 69.0 & 71.5 & 69.0 \\ 
Packages & 64.0 & 60.9 & 60.5 & 63.9 & 75.2 & 67.7 & 65.4 & 69.6 & 67.5 & 76.2 & 74.2 & 64.9 & 67.6 & 65.9 & 72.3 & 74.2 & 64.0 & 74.7 & 72.7 & 69.3 \\ 
Raccoon & 59.2 & 53.6 & 56.5 & 59.5 & 54.0 & 56.9 & 60.0 & 56.8 & 62.0 & 52.6 & 60.1 & 57.1 & 64.5 & 55.0 & 59.3 & 58.1 & 63.2 & 62.0 & 56.6 & 67.6 \\ 
NorthAmericaMushrooms & 76.3 & 60.4 & 38.4 & 65.6 & 59.9 & 84.5 & 73.0 & 87.7 & 64.1 & 75.4 & 74.6 & 78.1 & 88.4 & 85.7 & 68.2 & 88.1 & 72.8 & 88.1 & 83.6 & 83.7 \\ 
Pothole & 25.9 & 31.9 & 23.3 & 29.5 & 25.3 & 34.5 & 34.5 & 32.5 & 32.3 & 28.9 & 40.1 & 28.4 & 34.3 & 33.3 & 33.5 & 42.7 & 36.8 & 41.7 & 34.2 & 42.6 \\ 
OxfordPets(breed) & 11.1 & 13.7 & 7.9 & 13.4 & 11.8 & 26.4 & 29.6 & 21.4 & 28.6 & 23.9 & 38.3 & 38.6 & 39.6 & 40.5 & 37.8 & 56.8 & 60.9 & 58.3 & 60.1 & 56.0 \\
\end{tabular}
}
\end{table}
\renewcommand{\arraystretch}{1}

\renewcommand{\arraystretch}{1.2}
\begin{table}[t]
\footnotesize\centering
\caption{Detection performance for Faster RCNN. All methods are finetuned with Full-FT approach.}
\label{tbl:sup:all_results_fullft_frcnn}
\scalebox{0.55}{ 
\begin{tabular}{c||ccccc|ccccc|ccccc|ccccc}
\specialrule{0.8pt}{0pt}{0pt}
 & \multicolumn{5}{c|}{K=1} & \multicolumn{5}{c|}{K=3} & \multicolumn{5}{c|}{K=5} & \multicolumn{5}{c}{K=10} \\
Seed & 0 & 1 & 2 & 3 & 4 & 0 & 1 & 2 & 3 & 4 & 0 & 1 & 2 & 3 & 4 & 0 & 1 & 2 & 3 & 4 \\ \hline

CottontailRabbits & 44.0 & 16.5 & 41.9 & 36.4 & 48.4 & 56.9 & 41.9 & 57.0 & 45.1 & 49.3 & 54.7 & 59.3 & 43.3 & 55.2 & 54.8 & 62.7 & 38.5 & 47.0 & 39.1 & 53.2 \\ 
Packages & 11.1 & 59.9 & 40.4 & 44.0 & 64.3 & 46.6 & 60.6 & 66.3 & 55.7 & 65.3 & 36.0 & 67.7 & 45.6 & 49.5 & 55.2 & 62.1 & 70.3 & 48.1 & 63.1 & 67.6 \\ 
Raccoon & 30.2 & 20.5 & 17.5 & 51.3 & 26.2 & 31.8 & 38.6 & 7.0 & 45.2 & 30.6 & 47.4 & 38.5 & 13.9 & 40.6 & 30.7 & 33.3 & 47.6 & 35.5 & 44.6 & 35.4 \\ 
NorthAmericaMushrooms & 56.6 & 35.5 & 28.6 & 41.5 & 69.3 & 70.3 & 61.5 & 86.3 & 83.7 & 70.5 & 76.8 & 62.8 & 81.1 & 84.5 & 66.5 & 90.5 & 66.1 & 81.5 & 80.1 & 74.4 \\ 
Pothole & 3.8 & 15.0 & 3.5 & 6.3 & 10.5 & 19.7 & 18.1 & 12.7 & 27.3 & 8.3 & 22.3 & 21.7 & 20.1 & 27.3 & 13.8 & 26.9 & 21.7 & 26.7 & 33.0 & 26.0 \\ 
OxfordPets(breed) & 10.3 & 13.5 & 14.2 & 13.4 & 11.3 & 35.6 & 37.6 & 35.7 & 33.2 & 36.3 & 47.5 & 50.1 & 47.4 & 45.7 & 48.4 & 60.3 & 62.0 & 60.2 & 60.0 & 60.8 \\
\end{tabular}
}
\end{table}
\renewcommand{\arraystretch}{1}

\renewcommand{\arraystretch}{1.2}
\begin{table}[t]
\footnotesize\centering
\caption{Detection performance for F-ViT. All methods are finetuned with Full-FT approach.}
\label{tbl:sup:all_results_fullft_fvit}
\scalebox{0.55}{ 
\begin{tabular}{c||ccccc|ccccc|ccccc|ccccc}
\specialrule{0.8pt}{0pt}{0pt}
 & \multicolumn{5}{c|}{K=1} & \multicolumn{5}{c|}{K=3} & \multicolumn{5}{c|}{K=5} & \multicolumn{5}{c}{K=10} \\ 
Seed & 0 & 1 & 2 & 3 & 4 & 0 & 1 & 2 & 3 & 4 & 0 & 1 & 2 & 3 & 4 & 0 & 1 & 2 & 3 & 4 \\ \hline

CottontailRabbits & 67.2 & 68.4 & 66.3 & 68.8 & 64.7 & 65.8 & 67.0 & 64.6 & 64.7 & 70.5 & 67.7 & 67.9 & 72.8 & 70.1 & 68.8 & 66.4 & 74.6 & 64.7 & 69.4 & 64.3 \\ 
Packages & 67.0 & 60.6 & 58.3 & 48.0 & 67.2 & 59.7 & 57.0 & 67.6 & 53.4 & 68.2 & 65.4 & 63.1 & 59.9 & 63.9 & 62.0 & 68.6 & 70.2 & 70.9 & 63.4 & 68.6 \\ 
Raccoon & 54.9 & 61.5 & 62.5 & 57.8 & 57.6 & 49.0 & 55.8 & 57.5 & 46.8 & 59.4 & 49.8 & 56.3 & 58.0 & 49.6 & 56.9 & 40.1 & 53.7 & 53.6 & 54.7 & 58.9 \\ 
NorthAmericaMushrooms & 67.5 & 47.6 & 64.1 & 65.4 & 64.4 & 81.9 & 74.7 & 81.4 & 67.9 & 61.4 & 77.4 & 67.4 & 85.2 & 83.6 & 66.6 & 84.8 & 69.3 & 76.9 & 85.1 & 81.8 \\ 
Pothole & 10.6 & 8.4 & 12.8 & 13.3 & 20.0 & 19.6 & 16.3 & 19.1 & 23.6 & 19.3 & 26.7 & 15.7 & 27.7 & 26.8 & 26.8 & 25.9 & 28.7 & 31.1 & 30.5 & 27.7 \\ 
OxfordPets(breed) & 49.4 & 51.2 & 53.5 & 51.8 & 47.8 & 66.9 & 68.1 & 67.8 & 66.3 & 64.7 & 71.0 & 69.3 & 70.5 & 70.1 & 70.5 & 70.3 & 73.2 & 74.7 & 74.8 & 73.0 \\
\end{tabular}
}
\end{table}
\renewcommand{\arraystretch}{1}

%

\begin{figure}[t]
    \def\@captype{table}
    \begin{minipage}[t]{.48\textwidth}
    \begin{center}
    \tblcaption{Detection performance for DyHead with TFA.}
    \label{tbl:sup:all_results_tfa_dyhead}
    \scalebox{0.68}{
    \begin{tabular}{c||ccccc}
    \specialrule{0.8pt}{0pt}{0pt}
    Seed & 0 & 1 & 2 & 3 & 4 \\ \hline
    CottontailRabbits & 60.6 & 61.8 & 57.8 & 52.3 & 62.4 \\ 
    Packages & 51.9 & 65.4 & 70.2 & 61.8 & 75.6 \\ 
    Raccoon & 58.6 & 57.5 & 50.8 & 56.2 & 57.6 \\ 
    NorthAmericaMushrooms & 70.2 & 72.5 & 72.8 & 76.1 & 70.7 \\ 
    Pothole & 1.1 & 5.3 & 5.2 & 9.8 & 3.4 \\ 
    OxfordPets(breed) & 6.2 & 6.2 & 6.9 & 5.5 & 6.5 \\ 
    VehiclesOpenImages & 54.8 & 57.7 & 59.7 & 55.8 & 52.7 \\
\end{tabular}
    }
    \end{center}
    \end{minipage}
    \hfill
    \def\@captype{table}
    \begin{minipage}[t]{.48\textwidth}
    \begin{center}
    \tblcaption{Detection performance for GLIP(A) with TFA.}
    \label{tbl:sup:all_results_tfa_glipa}
    \scalebox{0.68}{
    \begin{tabular}{c||ccccc}
    \specialrule{0.8pt}{0pt}{0pt}
    Seed & 0 & 1 & 2 & 3 & 4 \\ \hline
    CottontailRabbits & 61.1 & 60.1 & 60.2 & 59.9 & 60.1 \\ 
    Packages & 54.0 & 62.3 & 64.3 & 60.3 & 66.6 \\ 
    Raccoon & 58.6 & 59.6 & 49.9 & 56.5 & 59.1 \\ 
    NorthAmericaMushrooms & 58.2 & 59.6 & 64.7 & 59.1 & 60.3 \\ 
    Pothole & 6.3 & 2.8 & 6.7 & 11.4 & 7.1 \\ 
    OxfordPets(breed) & 3.9 & 4.1 & 3.9 & 3.0 & 3.8 \\ 
    VehiclesOpenImages & 55.3 & 53.4 & 55.4 & 54.2 & 54.6 \\
\end{tabular}
    }
    \end{center}
    \end{minipage}
\end{figure}
\begin{figure}[t]
    \def\@captype{table}
    \begin{minipage}[t]{.48\textwidth}
    \begin{center}
    \tblcaption{Detection performance for Faster RCNN with TFA.}
    \label{tbl:sup:all_results_tfa_frcnn}
    \scalebox{0.68}{
    \begin{tabular}{c||ccccc}
    \specialrule{0.8pt}{0pt}{0pt}
    Seed & 0 & 1 & 2 & 3 & 4 \\ \hline
    CottontailRabbits & 55.8 & 38.9 & 37.2 & 36.5 & 46.2 \\ 
    Packages & 44.0 & 59.5 & 63.8 & 58.0 & 67.2 \\ 
    Raccoon & 50.4 & 54.6 & 54.4 & 46.6 & 54.1 \\ 
    NorthAmericaMushrooms & 58.0 & 56.1 & 54.4 & 50.5 & 60.8 \\ 
    Pothole & 1.7 & 4.9 & 1.4 & 6.0 & 3.6 \\ 
    OxfordPets(breed) & 7.9 & 7.7 & 7.6 & 6.7 & 8.6 \\ 
    VehiclesOpenImages & 47.8 & 51.3 & 49.0 & 51.7 & 49.8 \\
\end{tabular}
    }
    \end{center}
    \end{minipage}
    \hfill
    \def\@captype{table}
    \begin{minipage}[t]{.48\textwidth}
    \begin{center}
    \tblcaption{Detection performance for F-ViT with TFA.}
    \label{tbl:sup:all_results_tfa_fvit}
    \scalebox{0.68}{
    \begin{tabular}{c||ccccc}
    \specialrule{0.8pt}{0pt}{0pt}
    Seed & 0 & 1 & 2 & 3 & 4 \\ \hline
    CottontailRabbits & 42.9 & 41.6 & 51.2 & 49.5 & 50.8 \\ 
    Packages & 48.7 & 51.2 & 53.8 & 48.7 & 51.2 \\ 
    Raccoon & 56.1 & 54.3 & 58.3 & 55.5 & 57.3 \\ 
    NorthAmericaMushrooms & 21.7 & 15.6 & 15.0 & 24.2 & 19.2 \\ 
    Pothole & 0.2 & 0.0 & 0.0 & 0.0 & 0.1 \\ 
    OxfordPets(breed) & 2.7 & 2.5 & 2.6 & 2.6 & 2.6 \\ 
    VehiclesOpenImages & 31.1 & 35.6 & 33.5 & 34.6 & 32.0 \\
\end{tabular}
    }
    \end{center}
    \end{minipage}
\end{figure}

\begin{figure}[t]
    \def\@captype{table}
    \begin{minipage}[t]{.48\textwidth}
    \begin{center}
    \tblcaption{Detection performance for Faster RCNN with FSCE.}
    \label{tbl:sup:all_results_fsce_frcnn}
    \scalebox{0.65}{
    \begin{tabular}{c||ccccc}
    \specialrule{0.8pt}{0pt}{0pt}
    Seed & 0 & 1 & 2 & 3 & 4 \\ \hline
    CottontailRabbits & 45.2 & 61.5 & 62.8 & 56.1 & 58.9 \\ 
    Packages & 56.7 & 61.0 & 66.3 & 54.6 & 62.3 \\ 
    Raccoon & 53.5 & 50.8 & 43.5 & 48.1 & 56.0 \\ 
    NorthAmericaMushrooms & 73.5 & 58.6 & 69.2 & 84.2 & 59.2 \\ 
    Pothole & 2.7 & 10.1 & 3.6 & 13.3 & 4.1 \\ 
    OxfordPets(breed) & 28.8 & 33.6 & 26.8 & 24.6 & 30.6 \\ 
    VehiclesOpenImages & 46.4 & 49.0 & 44.3 & 46.5 & 40.6 \\
\end{tabular}
    }
    \end{center}
    \end{minipage}
    \hfill
    \def\@captype{table}
    \begin{minipage}[t]{.48\textwidth}
    \begin{center}
    \tblcaption{Detection performance for F-ViT with FSCE.}
    \label{tbl:sup:all_results_fsce_fvit}
    \scalebox{0.65}{
    \begin{tabular}{c||ccccc}
    \specialrule{0.8pt}{0pt}{0pt}
    Seed & 0 & 1 & 2 & 3 & 4 \\ \hline
    CottontailRabbits & 55.8 & 59.2 & 56.6 & 48.5 & 58.3 \\ 
    Packages & 60.0 & 55.8 & 64.3 & 48.8 & 58.4 \\ 
    Raccoon & 49.9 & 58.1 & 56.7 & 54.1 & 59.9 \\ 
    NorthAmericaMushrooms & 76.5 & 60.2 & 70.7 & 54.0 & 67.7 \\ 
    Pothole & 11.1 & 22.7 & 23.6 & 19.6 & 20.8 \\ 
    OxfordPets(breed) & 65.4 & 68.1 & 62.3 & 67.3 & 62.8 \\ 
    VehiclesOpenImages & 46.3 & 41.6 & 46.8 & 46.0 & 45.2 \\
\end{tabular}
    }
    \end{center}
    \end{minipage}
\end{figure}

\clearpage
\renewcommand{\arraystretch}{1.2}
\begin{table}[t]
\footnotesize\centering
\caption{All results for combined datasets with COCO and O365, used in Sec.~\ref{sec:results:pretrain_data_amount} in the main paper. All methods are trained with $K=3$ setting.}
\label{tbl:sup:all_results_data_amount}
\scalebox{0.58}{ 
\begin{tabular}{c||ccccc|ccccc|ccccc|ccccc}
\specialrule{0.8pt}{0pt}{0pt}
\#Images & \multicolumn{5}{c|}{2K (1\%)} & \multicolumn{5}{c|}{20K (10\%)} & \multicolumn{5}{c|}{0.10M (50\%)} & \multicolumn{5}{c}{0.20M (100\%)} \\ \hline
Seed & 0 & 1 & 2 & 3 & 4 & 0 & 1 & 2 & 3 & 4 & 0 & 1 & 2 & 3 & 4 & 0 & 1 & 2 & 3 & 4 \\ \hline

CottontailRabbits & 51.8 & 37.3 & 54.9 & 53.7 & 46.4 & 58.5 & 38.3 & 52.9 & 52.5 & 48.7 & 65.4 & 44.6 & 55.0 & 63.1 & 60.5 & 64.9 & 57.0 & 60.7 & 61.4 & 67.9 \\ 
Packages & 38.9 & 31.3 & 60.7 & 33.1 & 67.0 & 42.5 & 53.1 & 47.7 & 57.4 & 69.2 & 47.5 & 60.0 & 74.3 & 43.8 & 69.7 & 41.5 & 63.7 & 50.9 & 45.7 & 56.3 \\ 
Raccoon & 27.5 & 35.5 & 29.6 & 37.5 & 29.2 & 35.1 & 43.8 & 43.4 & 32.6 & 36.4 & 42.3 & 40.4 & 49.9 & 48.8 & 36.8 & 48.6 & 49.2 & 51.8 & 50.6 & 34.5 \\ 
NorthAmericaMushrooms & 82.6 & 61.7 & 82.8 & 80.7 & 66.5 & 83.1 & 67.4 & 81.5 & 82.1 & 68.9 & 78.6 & 63.2 & 85.0 & 85.9 & 70.5 & 81.4 & 62.4 & 85.7 & 84.5 & 64.0 \\ 
Pothole & 11.3 & 9.3 & 12.5 & 13.1 & 8.8 & 13.6 & 11.0 & 14.3 & 10.8 & 8.7 & 23.1 & 13.9 & 20.9 & 20.2 & 18.0 & 22.4 & 13.2 & 17.3 & 16.9 & 17.4 \\ 
OxfordPets(breed) & 24.9 & 27.6 & 26.9 & 28.3 & 27.6 & 29.6 & 31.5 & 28.2 & 28.1 & 29.7 & 29.4 & 32.2 & 35.2 & 36.7 & 34.3 & 35.5 & 35.5 & 33.1 & 34.2 & 32.7 \\
\end{tabular}
}
\end{table}
\renewcommand{\arraystretch}{1}
\end{document}